\documentclass[journal,twoside,web]{ieeecolor}
\usepackage{tmi}

\usepackage{cite}
\usepackage{amsmath,amssymb,amsfonts}
\usepackage{algorithmic}
\usepackage{graphicx}
\usepackage{textcomp}
\usepackage{algorithm}
\usepackage{multirow}
\usepackage{algorithmic}
\def\BibTeX{{\rm B\kern-.05em{\sc i\kern-.025em b}\kern-.08em
    T\kern-.1667em\lower.7ex\hbox{E}\kern-.125emX}}
\begin{document}
\title{Structure Embedded Nucleus Classification for Histopathology Images}
\author{Wei Lou, Xiang Wan, Guanbin Li, \IEEEmembership{Member, IEEE}, Xiaoying Lou, Chenghang Li, Feng Gao, and\\ Haofeng Li, \IEEEmembership{Member, IEEE}
\thanks{Corresponding author: Haofeng Li.}
\thanks{Wei Lou and Haofeng Li are with Shenzhen Research Institute of Big Data, Guangdong Provincial Key Laboratory of Big Data Computing, The Chinese University of Hong Kong at Shenzhen, Shenzhen 518172, China (e-mail: weilou@link.cuhk.edu.cn; lhaof@sribd.cn).}
\thanks{Xiang Wan is with Shenzhen Research Institute of Big Data, Guangdong Provincial Key Laboratory of Big Data Computing, The Chinese University of Hong Kong at Shenzhen, Shenzhen 518172, China, and also with Pazhou Lab, Guangzhou 510330, China (wanxiang@sribd.cn).}
\thanks{Guanbin Li is with the School of  Computer Science and Engineering, Sun Yat-sen University, Guangzhou 510006, China (e-mail: liguanbin@mail.sysu.edu.cn).}
\thanks{Xiaoying Lou is with Department of Pathology, Guangdong Provincial Key Laboratory of Colorectal and Pelvic Floor Diseases, the Sixth Affiliated Hospital of Sun Yat-sen University, Guangzhou, China (e-mail: louxy3@mail.sysu.edu.cn).}
\thanks{Chenghang Li is with Artificial Intelligence Thrust, The Hong Kong University of Science and Technology at Guangzhou, Guangzhou 510030, China (e-mail: cli136@connect.hkust-gz.edu.cn).}
\thanks{Feng Gao is with Department of Colorectal Surgery, Department of General Surgery, Guangdong Provincial Key Laboratory of Colorectal and Pelvic Floor Diseases, The Sixth Affiliated Hospital, Sun Yat-sen University, Guangzhou 510655, China, and also with Shanghai Artificial Intelligence Laboratory, Shanghai, China (e-mail: gaof57@mail.sysu.edu.cn).}
\thanks{This work was supported in part by Chinese Key-Area Research and Development Program of Guangdong Province (2020B0101350001), in part by the Natural Science Foundation of Guangdong Province of China (2023A1515011464), in part by the National Natural Science Foundation of China (No.62102267), the Guangdong Provincial Key Laboratory of Big Data Computing, The Chinese University of Hong Kong, Shenzhen, and National Key Clinical Discipline.}
}

\maketitle

\begin{abstract}
Nuclei classification provides valuable information for histopathology image analysis. However, the large variations in the appearance of different nuclei types cause difficulties in identifying nuclei. Most neural network based methods are affected by the local receptive field of convolutions, and pay less attention to the spatial distribution of nuclei or the irregular contour shape of a nucleus. 
In this paper, we first propose a novel polygon-structure feature learning mechanism that transforms a nucleus contour into a sequence of points sampled in order, and employ a recurrent neural network that aggregates the sequential change in distance between key points to obtain learnable shape features.
Next, we convert a histopathology image into a graph structure with nuclei as nodes, and build a graph neural network to embed the spatial distribution of nuclei into their representations. To capture the correlations between the categories of nuclei and their surrounding tissue patterns, we further introduce edge features that are defined as the background textures between adjacent nuclei. Lastly, we integrate both polygon and graph structure learning mechanisms into a whole framework that can extract intra and inter-nucleus structural characteristics for nuclei classification. Experimental results show that the proposed framework achieves significant improvements compared to the state-of-the-art methods.
\end{abstract}


\begin{figure}[!h]
\includegraphics[width=1.0\linewidth]{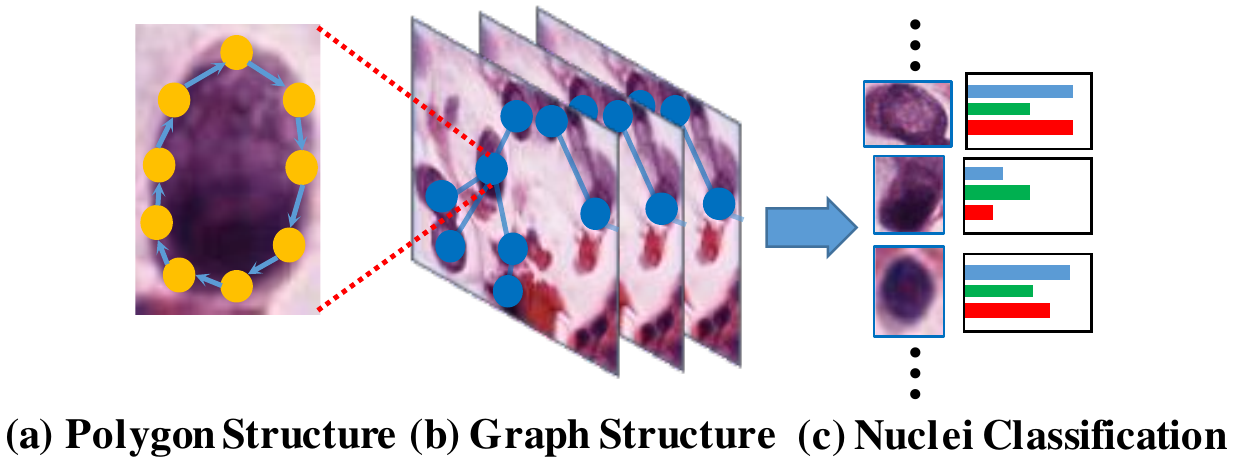}
\centering
\caption{Overview of the proposed nucleus classification framework that consists of (a) a polygon structure learning module, (b) a multi-layer GNN for graph structure learning, and (c) an instance-level classifier. }
\label{fig_1}
\end{figure}

\section{Introduction}
\label{sec:introduction}
\IEEEPARstart{R}{ecently}, computer-aided diagnosis (CAD) systems have achieved great success in histological examination tasks for their efficient, accurate and reproducible diagnosis performance. In a CAD system, nuclei segmentation and classification are critical steps for analyzing histopathology images. 
Segmenting a nucleus is to label all its pixels while nuclei classification is to identify the category of a nucleus. 
Solving these tasks not only obtains the size, texture and shape of each nucleus, but also provides the distribution among different nuclei types. 
The spatial relationships among different types of nuclei can effectively guide the CAD system in computational pathology tasks such as survival prediction~\cite{lu2018nuclear,tabibu2019pan}, cancer subtype classification and grading~\cite{ertosun2015automated,couture2018image}.

Nowadays, nuclei classification for histopathology images remains a challenge. Two nuclei of different types could have similar shapes and textures, and it is hard to distinguish between them. Meanwhile, for the nuclei of the same category, their appearances could have a wide variation during the different periods of their life cycle~\cite{abousamra2021multi,doan2022sonnet}. 
It is difficult for CAD systems or  inexperienced pathologists to classify every single nucleus in an image accurately. Thus, we aim at solving the nuclei classification task.

Most deep learning (DL) based methods~\cite{graham2019hover,abousamra2021multi,doan2022sonnet,ilyas2022tsfd} adopt Convolutional Neural Networks (CNNs) to detect and classify a nucleus based on its own convolutional features. However, these approaches seldom consider the relationship between different nuclei in an image. Such inter-nucleus information is crucial and widely used by experts in manual classification. 
Moreover, existing DL-based models rely on pixel-level representations. These approaches do not explicitly model the shape of a nucleus boundary, and ignore the structural information inside a nucleus. Recently, Graph Neural Networks (GNNs) based methods have grown rapidly in computational pathology~\cite{sureka2020visualization,zhao2020predicting,anklin2021learning,jaume2021quantifying}. Previous GNN models employ tissue patches or nuclei objects as graph nodes to construct a graph. However, most of them focus on classifying Whole Slide Images (WSIs) instead of nuclei. 

Inspired by the above observations, we propose to improve the nuclei classification by considering not only the correlations among different nuclei, but also the polygon structure of a nucleus shape.
First, to exploit the shape characteristics of a nucleus, we model a nucleus contour as a polygon, which can be described as an ordered sequence of points sampled from the contour (as shown in Fig.~\ref{fig_1}(a)). A recurrent neural network (RNN) is utilized to learn structural representations for the polygon from the point sequence, by aggregating the continuous changes in relative positions of sampled contour points. Such a polygon feature can well describe the irregular contour of a nucleus, and will act as a part of the nucleus representation.
Second, to harvest the spatial distribution among nuclei, we model a histopathology image as a graph by defining a node as a nucleus and an edge as the background between two adjacent nuclei. A graph neural network is exploited to capture inter-nuclei contexts via iteratively propagating each node information to its neighbors and updating each nucleus feature with its background and adjacent nodes.
The structural information of the whole histopathology graph is embedded into the enhanced representation of each nucleus. 
Lastly, we combine the above ideas of graph structure learning and polygon shape learning to develop a nuclei classification framework in which the pixel-level feature extraction and the nuclei-level classification are end-to-end trained.

Our main contributions are as follows:
\begin{itemize}
    \item A novel intra-nucleus polygon structure learning (PSL) module that learns the shape feature of a nucleus.
    \item A novel structure embedded nuclei classification framework based on an inter-nucleus graph structure learning (GSL) module and the proposed PSL module.
    \item The proposed framework significantly outperforms the state-of-the-art by 4.7\%-9.8\% average F-score on an in-house dataset and three public benchmarks. The experimental results show that both the proposed PSL and GSL modules can effectively improve the classification performance. 
\end{itemize}

\section{Related Work}

\subsection{Nuclei Classification for Histopathology Images}
In the early stage, handcrafted features of texture, morphology and color are extracted and sent into an SVM/AdaBoost classifier for nuclei classification~\cite{liu2011features,sharma2015multi}.
These methods explicitly model the intra-nucleus structure but are limited by the non-learnable representations.
Nowadays, most nuclei classifiers adopt CNNs with two stages, detecting nuclei instances and then labeling them~\cite{zhang2017deeppap,basha2018rccnet,zormpas2019superpixel,abousamra2021multi,graham2023one}.
Sirinukunwattana~\cite{sirinukunwattana2016locality} propose a CNN to detect nuclei centers and another CNN to classify the image patches containing a nucleus.
Graham~\cite{graham2019hover} proposes a CNN of three branches, two for segmentation, and one for classification. Doan~\cite{doan2022sonnet} predicts a weight map to highlight hard pixel samples for classification.
However, these approaches are limited by the receptive field of CNNs, and fail to harvest long-range contexts and spatial distributions of nuclei instances.


\subsection{Graph Models in Computational Pathology}
GNNs 
have become popular in computational pathology \cite{zhou2019cgc,ding2020feature,studer2021classification,hassan2022nucleus}. Most GNN methods are to classify a whole image~\cite{pati2022hierarchical} or tissue patches~\cite{javed2020cellular}. 
In these works, graph nodes are defined as tissue patches ~\cite{aygunecs2020graph}, nuclei objects \cite{jaume2021quantifying} or superpixels ~\cite{anklin2021learning}. The node embeddings can be hand-crafted~\cite{gunduz2004cell,zhou2019cgc} or extracted from pre-trained models~\cite{jaume2021quantifying,pati2022hierarchical}. 
Different from these works, we design a finer node representation including a shape feature of polygon-structure learning. 
Some existing work is based on GNN but it works by refining the nuclei classification results from existing models~\cite{hassan2022nucleus}. Instead, the GNN in our framework does not rely on nuclei types predicted in advance but directly performs the classification.


\subsection{Contour and Polygon Representation Learning}
Predicting polygons for object segmentation~\cite{castrejon2017annotating,schmidt2018cell,chen2023cpp} and contour detection~\cite{shen2015deepcontour} have been widely studied, while the feature extraction and classification for contours and polygons have been less discussed. Contour-aware nuclei segmentation methods~\cite{chen2016dcan} predict if a pixel is at a contour to improve the segmentation, but do not aggregate the pixel-level contour features to classify nuclei. PBC~\cite{huang2016pbc} is a polygon-based classifier that pools the texture features of each point in a polygon, but it does not consider the irregularity and smoothness of a polygon contour. Sharma\cite{sharma2015multi} computes morphological statistics (such as Area and Convexity of Contour) as the shape feature of a nucleus. Differently, we model a nucleus contour in a fine-grained way, using a sequence of relative positions between the centroid and vertices.


\begin{figure*}[!tbp]
\includegraphics[width=0.9\linewidth]{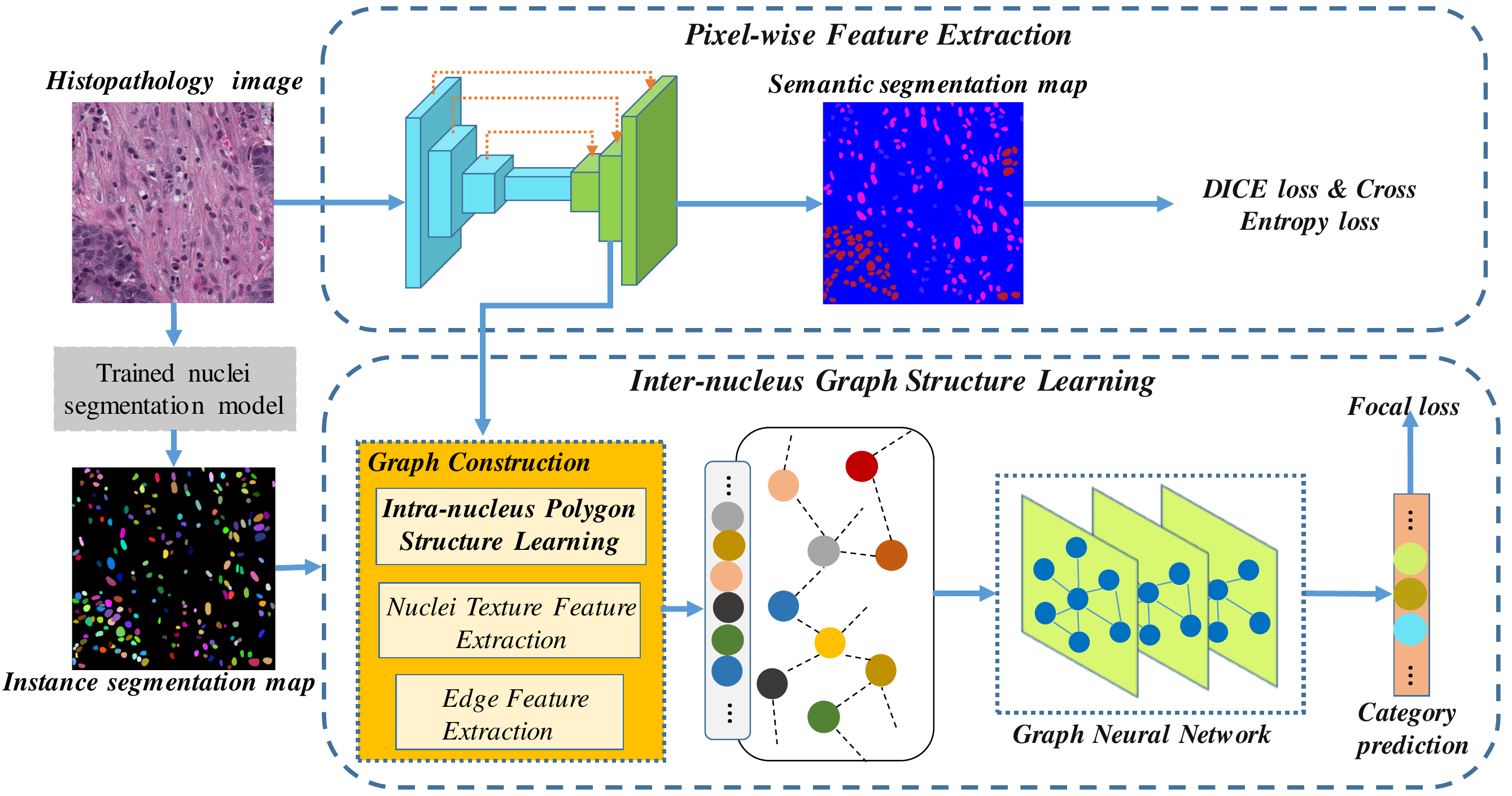}
\centering
\caption{Structure Embedded Nucleus Classification (SENC) framework. It consists of a pixel-wise feature extraction branch (upper) and an instance-level classification branch (lower) using the inter-nucleus Graph Structure Learning module (GSL). 
In the GSL module, an intra-nucleus Polygon Structure Learning module (PSL) computes the shape features for nuclei. Then the input image is transformed into a graph and a GNN enhances the features of the graph nodes for nuclei classification.
}
\label{fig_2}
\end{figure*}

\section{Methodology}

\subsection{Structure Embedded Nucleus Classification Framework}
The Structure Embedded Nuclei Classification (SENC) framework is illustrated in Fig.~\ref{fig_2}. The proposed framework consists of a pixel-wise feature extraction branch and an instance-level classification branch that adopts the modules of Inter-nucleus Graph Structure Learning (GSL) and Intra-nucleus Polygon Structure Learning (PSL). In the pixel-wise feature extraction branch, we utilize an existing CNN~\cite{guo2022visual} as the encoder and a feature pyramid network (FPN)~\cite{lin2017feature} as the decoder. The branch uses a histopathology image as input, and produces a pixel-level feature map from the second last decoder layer. In Fig~\ref{fig_2}, the semantic prediction map is the output of the decoder, when the encoder-decoder is pre-trained on a semantic segmentation task using DICE loss and cross-entropy loss.

The inter-nucleus GSL module takes a feature map and an instance map as inputs. The instance map is the result of nuclei instance segmentation. Since we focus on solving the classification task, we simply use an existing nuclei segmentation model~\cite{graham2019hover} to predict the instance map. The inter-nucleus GSL module first transforms a histopathology image into a graph structure, and then learns nuclei representation as graph nodes and background features as edges. In particular, we use a novel polygon structure learning module to obtain shape features for the nuclei nodes. After enhancing the node features with a graph neural network, all the nuclei are classified with a fully-connected (FC) layer and a Softmax layer.

\subsection{Inter-Nuclei Graph Structure Learning}
To capture the spatial relationship among nuclei, we develop an inter-nucleus graph structure learning module, which includes the construction of a histopathology graph topology, and the feature extraction of graph nodes and edges.

\textbf{Histopathology Graph Topology} Suppose a histopathology image contains $N$ nuclei. We construct a graph topology corresponding to the image by defining a graph node as a nucleus. The graph is undirected and can be defined as ${G}=\left({V}, {E}\right)$. $V$ denotes the set of nodes in the graph, namely, all the nuclei entities in the histopathology image. $E$ is the set of edges which represent the inner relationships between adjacent nuclei. To build an efficient and sparsely-connected graph, we first calculate the Euclidean distance between the centroid coordinates of any two nuclei. For each nucleus, $K$ undirected edges are linked between the nucleus and its $K$ nearest neighbors. 
The connection of nodes can be defined as a symmetric adjacency matrix $A \in R^{N \times N}$, where $A_{u, v}=1$ if the edge between the $u^{th}$ and $v^{th}$ nodes exists ($e_{u, v} \in E$). After computing the adjacency matrix, the original representations of nodes and edges, we use a GNN to update the node features and to harvest the structural embedded nuclei representation. The updated node features are exploited to predict the nuclei types.

\begin{figure*}[!t]
\includegraphics[width=0.95\linewidth]{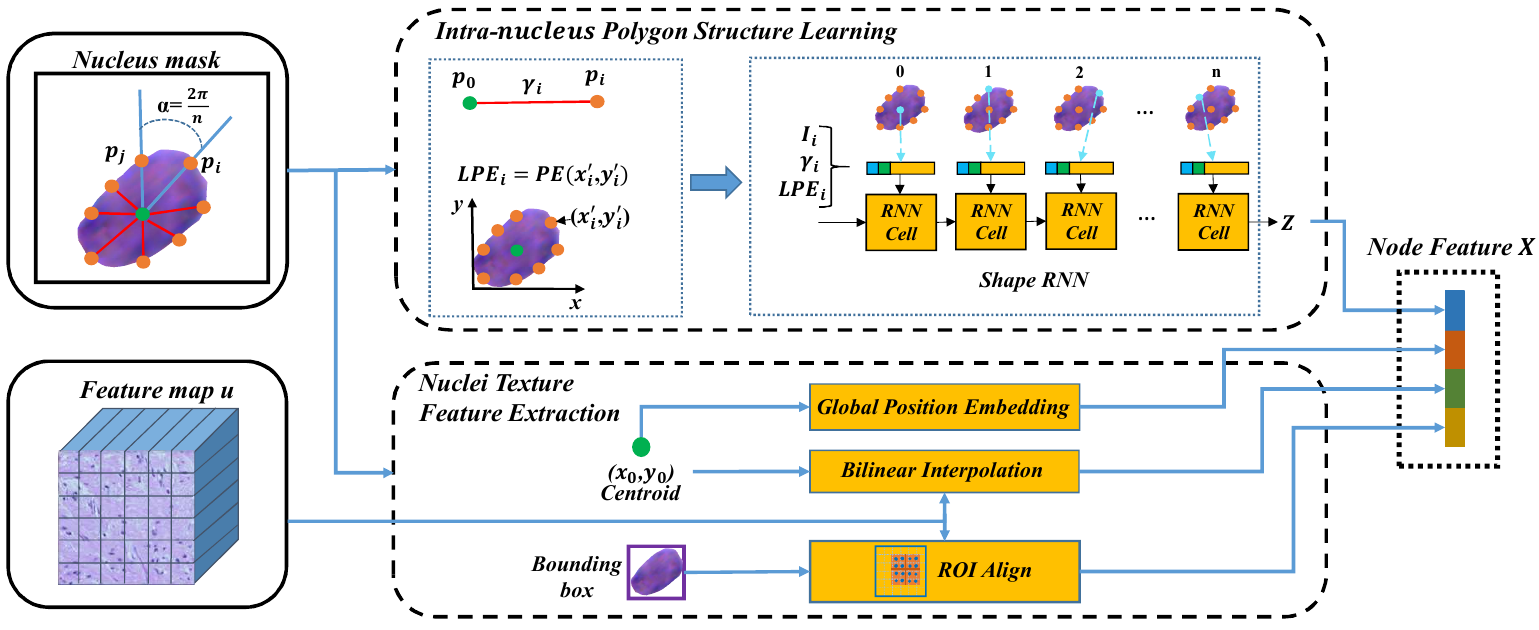}
\centering
\caption{Node feature extraction process for each nucleus. It consists of two parts: an intra-nucleus Polygon Structure Learning module (PSL) and a Nuclei Texture feature extraction module (NuTef). The PSL module uses a shape RNN to extract the feature $Z$ for a sequence of sampled contour points. $i/{\gamma_i}/ LPE_i$ denotes a point index, the Euclidean distances between $p_i$ and the centroid $p_0$, the position encoding of $p_i$ in a local coordinate system. 
The NuTef module computes the textures feature with the bounding box and centroid of a nucleus.
The feature map $u$ is produced by the pixel-wise feature extraction branch as shown in Fig.~\ref{fig_2}.}
\label{fig_3}
\end{figure*}

\textbf{Node feature extraction}
To obtain original node representations, we propose a node feature extraction process. The process takes the feature map from the pixel-weise feature extraction branch (Fig.~\ref{fig_2}) and a nucleus mask as inputs, and outputs a feature vector for the nucleus. The mask is extracted from an instance map predicted by a trained nuclei segmentation model. The shape of the feature map is $\frac{h}{4} \times \frac{w}{4} \times c$.
The node feature extraction includes an intra-nucleus Polygon Structure Learning module (PSL) and a Nucleus Texture feature extraction module (NuTef). The PSL module is to calculate the shape feature $Z$ for a nucleus, using a recurrent neural network (RNN). The PSL details are in the next subsection. The NuTef module is to learn the texture feature of a nucleus. As Fig.~\ref{fig_3} shows, ROI Align~\cite{he2017mask} is used with the input feature map to compute the representation $B \in R^{1 \times c}$ of the nucleus bounding box. The centroid coordinate ($h',w'$) of the nucleus is adopted to sample a feature vector $C \in R^{1 \times c}$ from the input feature map (of size $\frac{h}{4} \times \frac{w}{4} \times c$). In details, we locate the four pairs of integer coordinates nearest to ($h',w'$) and sample the centroid feature as the weighted combination of these four feature vectors, using the bilinear interpolation method. 
To introduce the information of the nucleus location in the original histopathology image, we compute a positional encoding vector $P E \in R^{1 \times c}$ for the nucleus centroid, using the Sinusoidal Position Encoding method~\cite{vaswani2017attention}. To obtain the final texture representation, the above three feature vectors are concatenated into a vector $T \in R^{1 \times 3c}$ as Eq.~(\ref{eq:texture}) shows:
\begin{align}\label{eq:texture}
T=concat(\left\{B, C, P E\right\}).
\end{align}
After computing the shape feature $Z$ with the PSL module, the feature vector of a node can be obtained by joining the texture and shape features as: $X = concat \left(T, Z\right)$.

\begin{figure}[!h]
\includegraphics[width=0.9\linewidth]{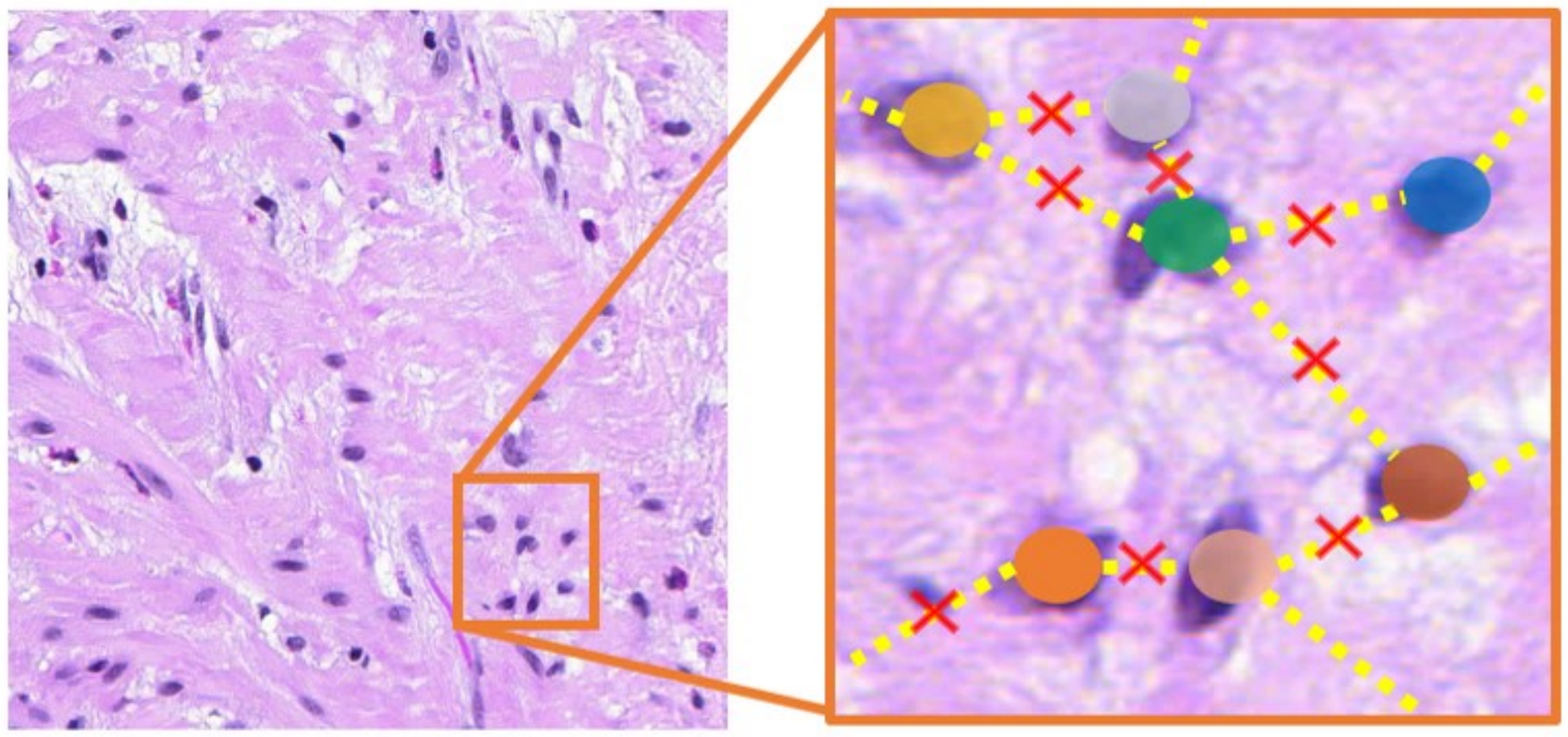}
\centering
\caption{Visualization of edge points for the feature extraction of edges. An edge point of two connected nuclei nodes is defined as the middle point (red cross) between the position of their centroids.}
\label{fig_4}
\end{figure}

\textbf{Edge feature extraction}
Edge features are widely used in GNNs to enhance the relationship reasoning between two neighboring nodes. In histopathology images, the category of a nucleus has correlations with its surrounding tissue backgrounds. 
Thus, we define a graph edge as the background regions between the two nuclei nodes of the edge. To extract an edge feature, we draw a line between two edge-connected nodes, and define an \textit{edge point} as the middle point (see the red crosses in Fig.~\ref{fig_4}) of the line. The edge feature is sampled from the pixel-wise feature map, using the coordinates of the edge point and the bilinear interpolation. 

\subsection{Intra-Nucleus Polygon Structure Learning} \label{sec:Intra-nuclei}
The intra-nucleus polygon structure learning module (PSL) is to model a nucleus contour as a polygon, and to learn the shape representation of the nucleus. For a nucleus, we compute its centroid position $p_{0}$ and sample $n$ points ${p_1,...,p_n}$ at the nucleus contour. To better describe the irregular shape, $n$ rays are emitted from the centroid with the same angle interval $\alpha=\frac{2 \pi}{n}$, and intersect the boundary at $n$ points that are collected in clockwise order to form a point sequence. To model the positions of contour points relative to the centroid, we insert the centroid at the head of the sequence.

We employ a multi-layer recurrent neural network (RNN) to harvest the shape feature with the point sequence. The input of the RNN is a feature sequence corresponding to the point sequence. The $i^{th}$ element of the feature sequence is defined as $S_{i}=\left(i, \gamma_{i}, L P E_{i}\right)$ to represent the initial feature of each sampled point $p_i$. $i$ denotes the index of the point sequence, $\gamma_{i}$ is the Euclidean distance between a sampled point $p_i$ and the centroid $p_0$. $LPE$ means the Local Position Encoding. $LPE_i$ denotes the position encoding vector of $p_i$, and is calculated in a local rectangular coordinate system, which is centered at the bottom-left corner of the bounding box of the nucleus. To run the RNN with the input sequence $S$, we set the hidden (input) states before the $1^{st}$ layer of the RNN as $h_{i}^{0}={S}_{i} (0\leq i\leq n)$. Then the hidden state of the $l^{th}$ layer and $i^{th}$ input position can be computed as Eq.~(\ref{eq:rnn}):
\begin{align}\label{eq:rnn}
h_{i}^{l}=\phi\left(h_{(i-1)}^{l} W^{l}_{s}+ h_{i}^{l-1} W^{l}_{h}\right),
\end{align}
where $W^{l}_{s}$ and $W^{l}_{h}$ are the weights of the input and recurrent neurons at the $l^{th}$ hidden layer, $h_{i-1}^{l}$ is the hidden state of the $(i-1)^{th}$ input position and the $(l)^{th}$ layer. $+$ denotes the elementwise addition. $\phi$ denotes the ReLU function. 
If $i-1$ is unavailable ($<0$), then $h_{(i-1)}^{l} W^{l}_{s}$ is ignored in Eq.~(\ref{eq:rnn}).
The RNN output is based on the hidden state of the final layer:
\begin{align}
Z=h_{n}^{M} W_{Z}, 
\end{align}
where $M$ is the layer number of the RNN. $W_Z$ is the weights in the output layer and $Z \in R^{1 \times c}$ is the output shape feature for the input nucleus node.

\subsection{Graph Neural Network Architecture}
Given a histopathology graph $G=\left({V}, {E} \right)$, its initial node features and edge features, we employ a GNN to harvest structure guided representations and to identify types for these nuclei nodes. A GNN usually consists of multiple layers. Each GNN layer aggregates and updates the node features from the previous layer or GNN input. In the aggregating step, the node features in a neighborhood are aggregated into a single feature via a differentiable operator. In the update step, each node feature is updated as the combination of its aggregated neighboring feature and itself. 

We implement the GNN using GENeralized Graph Convolution (GENConv)~\cite{li2020deepergcn} with the DeepGCN~\cite{li2021deepgcns} structure. GENConv is a graph convolution operator that can deal with edge features. Its key idea is to apply generalized mean-max aggregation functions by keeping the message features positive~\cite{li2020deepergcn}. The message aggregation and update processes can be formulated as Eq.~(\ref{eq4}) and Eq.~(\ref{eq5}), respectively:
\begin{gather}
a_{i}=\sigma\left(ReLU\left(X_j+ A_{i j} \cdot Y_{i j}\right)+\epsilon \right),j \in \Psi(i)\label{eq4}, \\ X_{i}=\zeta\left(X_i,a_{i}\right)\label{eq5}, 
\end{gather}
where $\sigma$ is the Softmax aggregations function~\cite{li2020deepergcn}, $X_i/X_j$ denotes the representation of the $i^{th}/j^{th}$ node, $\Psi(i)$ is the set of the neighbor indices of the $i^{th}$ node and $\epsilon$ is a small positive constant set to 1e-7. $A_{i j}$ is 1 if the $i^{th}$ and $j^{th}$ nodes are connected by an edge otherwise 0. $Y_{i j}$ is the feature of the edge between the $i^{th}$ and $j^{th}$ nodes. In the update function Eq.~(\ref{eq5}), $\zeta$ is a two-layer perceptron using ReLU as the activation functions, and is to enhance $X_i$ with the aggregated feature $a_i$.
To alleviate the vanishing gradient problem, we further utilize the residual connection following DeepGCN as shown in Eq.~(\ref{eq:res}):
\begin{align}\label{eq:res}
X^{l+1}=\mathcal{F}\left(X^{l}\right)+X^{l}, 
\end{align}
where $\mathcal{F}(\cdot)$ contains a GENConv layer, a batch normalization layer and a ReLU activation function. $X^{l}$ denotes all the node features $X_i^{l}$ produced by the $l^{th}$ GNN layer. Finally, with $X^L$ the node representations updated by the last GNN layer, all the nodes are predicted through a classifier described in Eq.~(\ref{eq:cls}):
\begin{align}\label{eq:cls}
t=Softmax\left(FC\left(X^{L}\right)\right). 
\end{align}
\subsection{Training Scheme}
In the proposed framework, the pixel-wise feature extraction branch and the instance-level classification branch are trained simultaneously in an end-to-end manner. The feature extraction branch aims to learn rich texture features by solving a semantic segmentation task with the Dice loss in Eq.~(\ref{eq8}) and the Cross-entropy loss in Eq.~(\ref{ep9}):
\begin{align}
\mathcal{L}_{\text{CE}}=-\frac{1}{H\times W} \sum_{i=1}^{H \times W} \sum_{q=1}^{Q} y_{i, q}^s \log x_{i, q}^s, \label{eq8}
\end{align}
\begin{align}
\mathcal{L}_{\text{Dice}} = 1-\frac{2 \times \sum_{i=1}^{H \times W}\sum_{q=1}^{Q}\left( y_{i,q}^s \times x_{i,q}^s \right)+\varepsilon}{\sum_{i=1}^{H \times W}\sum_{q=1}^{Q} (y_{i,q}^s+x_{i,q}^s)+\varepsilon}, \label{ep9}
\end{align}
where $x^s$ is a $HW\times Q$ predicted map of the semantic segmentation task and $y^s$ is the ground truth map. $Q$ is the number of nuclei types, $H/W$ denotes the height/width of a GT or predicted map. $\varepsilon$ is a smoothness constant set to 1e-8.

In the instance-level classification branch using the PSL \& GSL modules, the classifier predicts category-wise probabilities for each nucleus entity. Cross-entropy loss is widely used in multi-class classification tasks. 
However, in the nuclei classification task, the distribution of different categories is usually unbalanced and easy samples (such as the nuclei of some tumor cells) could be too dominant to train a robust model.
Therefore, we utilize a Focal loss~\cite{lin2017focal} to pay more attentions to hard samples as Eq.~(\ref{eq:focal}) shows:
\begin{align}\label{eq:focal}
\mathcal{L}_{\text{Focal}}=-\frac{1}{N} \sum_{i=1}^{N} \sum_{q=1}^{Q} \tau_{q} (1-t_{i, q})^\gamma y_{i, q}^o \log t_{i, q}, 
\end{align}
where $t$ ($\in R^{N\times Q}$) contains the predicted probabilities for $N$ nuclei objects, $y^o$ is the true labels. $\gamma$ is a hyper-parameter to make the network concentrate on hard samples. The higher the value of $\gamma$, the lower the loss for well-classified examples. $\tau_{q}$ is the weight for each category and is set to the reciprocal of the proportion of the $q^{th}$ class in the training set.
The overall objective to train the network is described as Eq.~(\ref{eq:loss}):
\begin{align} \label{eq:loss}
\mathcal{L}\left(x^{s},t\right)=\mathcal{L}_{\text{Dice}}\left(x^s\right)+\mathcal{L}_{\text{CE}}\left(x^s\right)+\mathcal{L}_{\text{Focal}}\left(t\right), 
\end{align}
which is composed of two semantic segmentation losses (Dice \& Cross-entropy loss) and a classification loss (Focal loss). All three losses are equally weighted.

\begin{table*}[!htb]
\centering
\setlength{\tabcolsep}{2.0pt}
\caption{Quantitative comparison between existing nuclei segmentation \& classification models without and with our method (+Ours) on the CoNSeP, MoNuSAC, CRC-FFPE and PanNuke datasets. `$\textit{Net}$+Ours' means our classification method taking the segmentation results of the $\textit{Net}$ as input. 
$F^m_{c}$,$F^i_c$,$F^e_c$,$F^s_c$, $F^l_c$,  $F^{ma}_c$,  $F^n_c$, $F^t_{c}$, $F^{st}_{c}$,$F^{im}_{c}$,$F^{ne}_{c}$,$F^o_{c}$,$F^c_{c}$,$F^d_{c}$,$F^{ne}_{c}$  represent the F-score for the nuclei types of miscellaneous, inflammatory, epithelial, spindle-shaped, lymphocytes, macrophages, neutrophils, tumor, stroma, immune, necrosis, other, connective, dead, neoplastic, respectively. $F_{avg}$ is the average F-score of the categories in a dataset. `Imp.' is the classification improvement when using our framework.}
\begin{tabular}{cc|ccccccccc|cccccccccc}
\multirow{2}{*}{}& \multirow{2}{*}{Method} & \multicolumn{9}{c|}{CoNSeP} & \multicolumn{10}{c}{CRC-FFPE} \\ \cline{3-21}
& & $AJI$ & $PQ$ & $F_d$& $F_{avg}$ & $Imp$ &{$F^m_{c}$}  & {$F^i_c$}  & {$F^e_c$}  & {$F^s_c$}  & $AJI$ & $PQ$ & $F_d$& $F_{avg}$ & $Imp.$ & $F^{i}_{c}$&$F^{c}_{c}$&$F^{d}_{c}$&$F^{ep}_{c}$&$F^{ne}_{c}$   \\ \hline 
& MCSpatNet~\cite{abousamra2021multi} & - & - & 0.733 & 0.514 & - & 0.400 & 0.537 & 0.582 & 0.540 & -  & - & 0.752 & 0.278 & - &  0.547 & 0.196 & 0.307 & 0.126 & 0.212   \\

& Triple U-net~\cite{zhao2020triple} & 0.453 &0.412 &0.663&0.383& - & 0.102&0.570&0.423&0.438&0.510 &0.547 &0.684&0.196& - & 0.458&0.101&0.212&0.079&0.129 \\
& +Ours &  0.453 &0.412 &0.663 & 0.421 & +3.8\% & 0.231 & 0.632 & 0.476 & 0.478  &0.510 &0.547 &0.684 & 0.255 & +5.9\% & 0.479 & 0.134 & 0.320 & 0.111 & 0.231\\

& Mask2former~\cite{cheng2022masked}  & 0.464 & 0.482 & 0.659 & 0.414 &- & 0.325   & 0.461 & 0.462 & 0.408 & 0.451 & 0.491 & 0.581 & 0.212 &- & 0.299   & 0.164 & 0.278 & 0.101 &0.218 \\
& +Ours  & 0.464 & 0.482 & 0.659 & 0.536 & +12.2\% &0.501   & 0.548 & 0.561 & 0.532 & 0.451 & 0.491 & 0.581 & 0.252 & +4\% & 0.399   & 0.168 & 0.312 & 0.121 & 0.263 \\

& TSFD-net~\cite{ilyas2022tsfd}  & 0.458 & 0.415 & 0.680 & 0.439 &- & 0.120   & 0.567 & 0.558 & 0.513 & 0.478 & 0.503 & 0.549 & 0.182 & - & 0.308   & 0.110 & 0.201 & 0.098 &0.194 \\
& +Ours  & 0.458 & 0.415 & 0.680 & 0.548 &+10.9\% & 0.462   & 0.603 & 0.575 & 0.554 & 0.478 & 0.503 & 0.549 & 0.217 &+2.5\% & 0.341   & 0.145 & 0.278 & 0.101 & 0.223  \\

& HoVer-Net~\cite{graham2019hover} & 0.544 & 0.510 & 0.740 & 0.549  & - & 0.430 & 0.601 & 0.612 & 0.552 & 0.603 & 0.636 & 0.743 & 0.245 &- & 0.512   & 0.180 & 0.290 & 0.096 & 0.148  \\
& +Ours & 0.544 & 0.510 & 0.740 & \textbf{0.595} & +4.6\% & \textbf{0.510} & \textbf{0.632} & \textbf{0.646} & \textbf{0.592} & 0.603 & 0.636 & 0.743&\textbf{0.353}& +10.8\% & \textbf{0.548}&\textbf{0.196}&\textbf{0.418}&\textbf{0.159}&\textbf{0.446} \\
\hline
\multirow{2}{*}{}& \multirow{2}{*}{Method} & \multicolumn{9}{c|}{MoNuSAC} & \multicolumn{10}{c}{PanNuke} \\ \cline{3-21}
& & $AJI$ & $PQ$ & $F_d$& $ F_{avg}$ & $Imp$ & {$F^e_c$}  & {$F^l_c$}  &{$F^{ma}_c$}  & {$F^n_c$}  &$AJI$ & $PQ$ & $F_d$& $F_{avg}$ & $Imp.$ & $F^{i}_{c}$&$F^{c}_{c}$&$F^{d}_{c}$&$F^{ep}_{c}$&$F^{ne}_{c}$  \\ \hline
& MCSpatNet~\cite{abousamra2021multi} &  - & - & 0.828 & 0.651 & - & 0.698 & 0.753 & \textbf{0.517} & 0.636  & - & - & 0.786 & 0.483 &-& 0.484  & 0.473 & 0.220 & 0.612 & 0.629   \\

& Triple U-net~\cite{zhao2020triple} & 0.408&0.543&0.638&0.464&- &0.557&0.637&0.276&0.386 &0.583&0.464&0.698&0.381&- &0.392&0.298&\textbf{0.297}&0.508&0.412 \\
& +Ours &   0.408&0.543&0.638 & 0.536 & +7.2\% & 0.599 & 0.768 & 0.302 & 0.477  & 0.583 & 0.464  & 0.698 & 0.428 &+4.7\% & 0.436 & 0.361 & 0.324 & 0.564 & 0.457 \\

& TSFD-net~\cite{ilyas2022tsfd}  & 0.461 & 0.499 & 0.735 & 0.350 &- & 0.544 & 0.738 & 0.091 & 0.027  & 0.621 & 0.513 & 0.748 & 0.413 &- & 0.441 & 0.453 & 0.102 & 0.505 & 0.565 \\
& +Ours  & 0.461 & 0.499 & 0.735 & 0.435 & +8.5\% & 0.621 & 0.788 & 0.210 & 0.124  &  0.621 & 0.513 & 0.748 & 0.445 & +3.2\% &0.482  & 0.477 & 0.122 & 0.545 & 0.601 \\

& Mask2former~\cite{cheng2022masked}  & 0.577 & 0.568 & 0.746 & 0.559 &- & 0.682 & 0.762 & 0.349 & 0.446 &  0.616 & 0.666 & 0.792 & 0.480 &-& 0.400   & 0.426 & 0.289 & 0.668 &0.617 \\
& +Ours  &  0.577 & 0.568 & 0.746 & 0.603 & +4.4\% & 0.747 & 0.786 & 0.377 & 0.502 & 0.616 & 0.666 & 0.792 & 0.519 & +3.9\% & 0.514 & 0.497 & 0.256 & \textbf{0.688} & 0.642 \\

& HoVer-Net~\cite{graham2019hover} & 0.599 & 0.613 & 0.748 & 0.604 & - &0.754 & 0.806 & 0.394 & 0.463  & 0.653   & 0.621 &0.787 & 0.503 &- &0.530 & 0.474 & 0.260 & 0.618 & 0.632   \\
& +Ours &  0.599 & 0.613 & 0.748 & \textbf{0.692} &+8.8\% & \textbf{0.799} & \textbf{0.846} & 0.383 & \textbf{0.742} & 0.653   & 0.621 &0.787 & \textbf{0.528}& +2.5\% &\textbf{0.541}& \textbf{0.507}&0.263&0.684& \textbf{0.643} \\
\end{tabular}

\label{tab_SOTA}
\end{table*}

\section{Experiments}
\subsection{Datasets}
The proposed framework is evaluated on four nuclei classification datasets: CRC-FFPE, CoNSeP~\cite{graham2019hover}, PanNuke~\cite{gamper2020pannuke}, MoNuSAC~\cite{verma2021monusac2020}. The CRC-FFPE dataset is an in-house colorectal cancer dataset that consists of 16 patients with 59 H\&E stained histopathology tiles of size $1000 \times 1000$. The images are extracted from the WSIs collected from TCGA~\cite{weinstein2013cancer} and annotated by the pathologists in a local hospital. The nuclei types of the CRC-FFPE dataset include Tumor, Stroma, Immune, Necrosis, and Other. These images are divided into a training set (45 tiles) and a testing set (14 tiles).
The CoNSep dataset is a colorectal adenocarcinoma dataset that contains 41 H\&E stained images of size $1000 \times 1000$. The dataset includes 24139 annotated nuclei that are grouped into four categories: Miscellaneous, Inflammatory, Epithelial, and Spindle-shaped. We split the CoNSeP dataset into a training set with 27 images and a testing set with 14 images. The MoNuSAC dataset is a multi-organ dataset, comprising 310 images (209 for training, 101 for testing) of 71 patients. The size of images ranges from $81\times113$ pixels to $1422 \times 2162$ pixels. The dataset contains four types of organs (breast, kidney, lung, and prostate). The nuclei types of the dataset are: Epithelial, Lymphocytes, Macrophages, and Neutrophils. The PanNuke dataset contains 7899 image tiles of size $256 \times 256$ of 19 different organs. The images were digitized at 20$\times$ or 40$\times$ magnification. The nuclei types of the dataset are Inflammatory, Connective, Dead, Epithelial and Neoplastic.

\subsection{Implementation Details}
For the CRC-FFPE and CoNSeP datasets, all the training images are resized to $1024 \times 1024$. For the MoNuSAC dataset, we crop image patches of size $512 \times 512$. For the PanNuke dataset, we set the original image size to $256 \times 256$. We implement the proposed framework with PyTorch~\cite{paszke2017automatic} and PyTorch Geometric library~\cite{fey2019fast}. The encoder in the pixel-wise extraction branch consists of 4 layers with kernel sizes [3,3,12,3] and channel sizes [64,128,320,512], following the previous work~\cite{guo2022visual}. The encoder is pre-trained on ImageNet~\cite{deng2009imagenet}. For the GSL module, the GCN model comprises two GENConv~\cite{li2020deepergcn} layers ($L$=2) of 64 hidden channels. The neighbor number $K$ is set to 4 for building edges of a graph.
For the proposed PSL module, the channel number $c$ of each feature vector for the bounding box, centroid, and positional embedding is 64. The number of hidden layers $M$ of the RNN model is 2 and each layer has 128 hidden units. The number of sampled contour points $n$ in PSL is set to 18. The proposed framework is trained for 100 epochs with 
an Adam optimizer, an initial learning rate of $1\times10^{-4}$, and a momentum of 0.9 and 0.99. For all four datasets, $\gamma$ in the Focal loss is set to 2. 


\begin{table*}[!t]
\centering
\setlength{\tabcolsep}{2.8pt}
\caption{Quantitative comparison between existing methods and ours for nuclei classification on the CoNSeP, MoNuSAC, CRC-FFPE and PanNuke datasets in the `Inst-gt' setting. The ground truth of nuclei instance segmentation is accessible for all the methods in the inference stage.}
\begin{tabular}{cc|ccccc|cccccc}
\multirow{2}{*}{}& \multirow{2}{*}{Method} & \multicolumn{5}{c|}{CoNSeP} & \multicolumn{6}{c}{CRC-FFPE} \\ \cline{3-13}
& & $F_{avg}$ & {$F^m_{c}$}  & {$F^i_c$}  & {$F^e_c$}  & {$F^s_c$}    & $ F_{avg}$ & $F^t_{c}$ & $F^{st}_{c}$ & $F^{im}_{c}$ & $F^{ne}_{c}$ & $F^o_{c}$ \\ \hline 
& HoVer-Net~\cite{graham2019hover}  & 0.711 & 0.585 & 0.656 & 0.872 & 0.730 & 0.350 & 0.737 & 0.240 & 0.290   & 0.153 & 0.327  \\
& Triple U-net~\cite{zhao2020triple}  &0.438 &0.170&0.399&0.684&0.501 &0.289 &0.599&0.202&0.344&0.099&0.202 \\

& TSFD-net~\cite{ilyas2022tsfd}  & 0.632 & 0.502 & 0.635 & 0.822   & 0.570   & 0.221 & 0.378 & 0.176 & 0.222   & 0.100 & 0.233 \\
& MCSpatNet~\cite{abousamra2021multi}  & 0.695 & 0.562 & 0.624 & 0.863 & 0.730  & 0.341 & 0.772& 0.190 & 0.372 & 0.071 & 0.301   \\
& Ours & \textbf{0.777} & \textbf{0.726} & \textbf{0.685} & \textbf{0.890}   & \textbf{0.807} & \textbf{0.448} & \textbf{0.804} & \textbf{0.271} & \textbf{0.449}   & \textbf{0.184} & \textbf{0.534}  \\ \hline 

\multirow{2}{*}{}& \multirow{2}{*}{Method} & \multicolumn{5}{c|}{MoNuSAC} & \multicolumn{6}{c}{PanNuke} \\ \cline{3-13}
& & $F_{avg}$ &{$F^e_c$}  & {$F^l_c$}  &{$F^{ma}_c$}  & {$F^n_c$}  & $F_{avg}$&$F^{i}_{c}$&$F^{c}_{c}$&$F^{d}_{c}$&$F^{ep}_{c}$&$F^{ne}_{c}$ \\ \hline
& HoVer-Net~\cite{graham2019hover}  & 0.701 & 0.794  & 0.910 & 0.612 & 0.486   &  0.529 & 0.525 & 0.610 & 0.278 & 0.636 & 0.598  \\

& Triple U-net~\cite{zhao2020triple} &0.530&0.629&0.747&0.331&0.412&0.396 &0.412&0.278&0.304&0.555&0.432 \\

& TSFD-net~\cite{ilyas2022tsfd}   & 0.605 & 0.695 & 0.902 & 0.435 & 0.388  &0.418& 0.423 & 0.210 & \textbf{0.378} & 0.602 & 0.478 \\

& MCSpatNet~\cite{abousamra2021multi}  & 0.828 & 0.955 & 0.963 & 0.683 & 0.712  & 0.574 & 0.549 & 0.576 & 0.269 & 0.702 & 0.772   \\
& Ours   & \textbf{0.919} & \textbf{0.967} &\textbf{0.972} & \textbf{0.862}& \textbf{0.873} & \textbf{0.621} & \textbf{0.607} & \textbf{0.622} & 0.253 & \textbf{0.823} & \textbf{0.799} \\
\end{tabular}

\label{tab_SOTA_gt}
\end{table*}

\begin{figure*}[!t]
\includegraphics[width=0.99\linewidth]{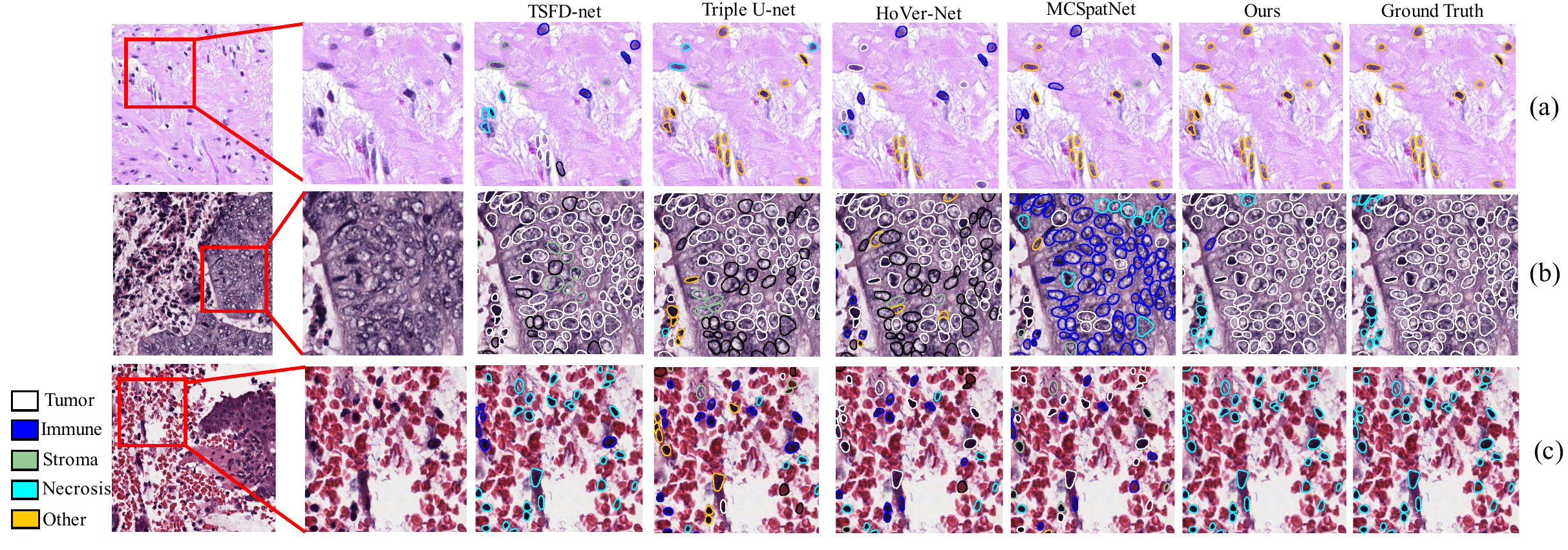}
\centering
\caption{Visualized classification results on the CRC-FFPE dataset and the 'Inst-gt' setting where the GTs of nuclei instance segmentation are accessible. We compare the classification performance between our framework and the existing methods. 
}
\label{fig_visual}
\end{figure*}

\subsection{Evaluation}
Following the previous works~\cite{graham2019hover,abousamra2021multi,doan2022sonnet}, we use an F-score $F_c$ to evaluate the nuclei classification methods. The F-score considers the performance of both detection and classification. 
Given a set of predicted nuclei and a set of ground truth (GT) nuclei, we assign each GT nucleus with its nearest predicted nucleus if their centroids are within 12 pixels, and ensure that no two GT nuclei are assigned to the same predicted nucleus. The predicted nuclei then can be split into the detected, undetected, and wrongly detected ones, whose numbers are denoted as ${TP}_{d}$, ${FN}_d$, ${FP}_{d}$, respectively. 
The classification performance is measured based on ${TP}_{d}$. For one of the categories (for example, the type $q$), the number of correctly classified nuclei, wrongly classified nuclei, correctly classified nuclei of types other than $q$, and wrongly classified nuclei instances of types other than type $q$ are denoted as $TP_{c}$, $FP_{c}$, $TN_{c}$ and $FN_{c}$, respectively. 
The F-score of type $q$ is computed as Eq.~(\ref{eq:fscore}):
\begin{align}\label{eq:fscore}
F_{c}^{q}=\frac{2\left(T P_{c}+T N_{c}\right)}{\left[\begin{array}{c}2\left(T P_{c}+T N_{c}\right)+\alpha_{0} F P_{c}+\alpha_{1} F N_{c} \\ +\alpha_{2} F P_{d}+\alpha_{3} F N_{d}\end{array}\right]}, 
\end{align}
where $\alpha_{1}=2$, $\alpha_{2}=2$, $\alpha_{3}=1$ and $\alpha_{4}=1$. The average F-score of all categories in a dataset is reported as $F_{avg}$ and can be viewed as a general metric of classification performance.

To show the detection and segmentation results of previous works, we evaluate the nuclei segmentation metrics: Aggregated Jaccard Index
(AJI)~\cite{Mahmood2019Deep}, Panoptic Quality (PQ)~\cite{kirillov2019panoptic}, Detection Quality $F_d$~\cite{graham2019hover}. AJI is an extension of the global Jaccard index and measures the overlapping areas of multiple objects and is recognized as an object-level criterion for segmentation evaluation. PQ is another metric for accurate quantification of detection and segmentation. It is defined as $\mathcal{P} \mathcal{Q}=\frac{|T P|}{|T P|+\frac{1}{2}|F P|+\frac{1}{2}|F N|} \times \frac{\sum_{(x, y) \in T P} \operatorname{IoU}(x, y)}{|T P|}$. The first part of PQ is the detection quality $F_d$. Each prediction-GT pairs are matched to be unique if their IoU(x,y) is larger than 0.5. The predictions and GT are split into matched pairs (TP), unmatched GT (FN) and unmatched predictions (FP). The detection quality $F_d$ then is defined as the $F_1$ score for instance detection. The second part of PQ is the segmentation quality which can be interpreted as how close each correctly detected instance is to its matched GT.

\subsection{Comparison with the state-of-the-arts}
We compare our proposed approach with the state-of-the-art nuclei classification methods HoVer-Net~\cite{graham2019hover}, Triple U-net~\cite{zhao2020triple}, MCSpatNet~\cite{abousamra2021multi}, TSFD-net~\cite{ilyas2022tsfd}, and Mask2former~\cite{cheng2022masked}. Among them, HoVer-Net, MCSpatNet, and Mask2former support nuclei classification, 
TSFD-net is a semantic segmentation method and Triple U-net is an instance segmentation method. For HoVer-Net, MCSpatNet and Mask2former, we directly compare the instance classification performance using their original settings. For TSFD-net, we extract the instance classification results using its semantic segmentation outputs and instance segmentation outputs. For Triple U-net, an extra $1 \times 1$ convolution is added as the classification layer.
Note that we aim at solving the classification task in this paper. To fairly compare the classification performance among existing methods and ours, we propose an evaluation setting `Inst-gt' where the GT maps of nuclei instance segmentation are accessible to all these methods during the testing stage. In this setting, these methods use the GT segmentation maps to replace their own predicted segmentation results for classifying nuclei. Since each method adopts the segmentation results of the same quality, the `Inst-gt' setting provides a fair comparison of nuclei classification. We also report the results on a typical setting `Inst-pred' where the segmentation GTs are not accessible and each method needs to predict its own segmentation results.
%

As TABLE~\ref{tab_SOTA} shows, in the 'Inst-pred' setting the proposed framework outperforms the state-of-the-art methods on all 4 types in the CoNSeP dataset, all 5 types in the CRC-FFPE dataset, 3 types in the MoNuSAC dataset and 4 types in the PanNuke dataset. 
`\textit{Model} + Ours' denotes our proposed framework taking the segmentation predictions of the \textit{Model} as input. 
`Imp.' is the classification F1-score improvement over each previous method when using our classification framework. The results display that our method can significantly improve the previous nuclei classification methods  by 3.8\%-12.2\%, 4.4\-8.8\%, 2.5\%-10.8\%, 2.5\%-3.9\%
 average F-score on the CoNSeP, MoNuSAC, CRC-FFPE and PanNuke datasets, respectively. In TABLE~\ref{tab_SOTA_gt}, on the 'Inst-gt' setting, our proposed framework achieves the highest F-score on all 4 types in the CoNSeP dataset, 
all 4 types in the MoNuSAC dataset 
, all 5 types in the CRC-FFPE dataset and 4 types in the PanNuke dataset. 
The proposed method outperforms the second-best model by 6.6\%, 9.1\%, 9.8\%, and 4.7\% average F-score on the four datasets. The results indicate that our proposed framework has achieved the state-of-the-art performance for nuclei classification in histopathology images.

Fig.~\ref{fig_visual} visualizes the nuclei classification results of some existing methods and ours on the CRC-FFPE dataset. The visual results are obtained on the 'Inst-gt' setting to compare only the classification performance among these approaches. 
In Fig.~\ref{fig_visual}(a), our method can accurately identify sparsely-distributed nuclei, since a GCN is utilized to propagate contextual information among even remote nuclei. In Fig.~\ref{fig_visual}(b), the proposed framework shows its advantage of classifying densely-distributed nuclei of the same type. Most of these nuclei are surrounded by similar background regions, which can be well described by the edge representations and guide the nuclei classification in our proposed method.


\begin{table}[!htb]
\centering
\caption{Ablation study on the Consep dataset and using the ground truth as segmentation map in the inference stage.
}
\begin{tabular}{c|ccccc}
\multirow{2}{*}{Method}   & \multicolumn{5}{c}{CoNSeP} \\ \cline{2-6}
 & $F_{avg}$ & {$F^m_{c}$}  & {$F^i_c$}  & {$F^e_c$}  & {$F^s_c$}\\ \hline 
(a) Baseline       & 0.437 & 0.187 & 0.416  & 0.721 & 0.422   \\
(b) +GCN          & 0.651 & 0.437 & 0.575  & 0.869 & 0.714   \\
(c) +EdgeFeat  & 0.690 & 0.580 & 0.592 & 0.879 & 0.704   \\
(d) Ours     & \textbf{0.777} & \textbf{0.726} & \textbf{0.685} & \textbf{0.890} & \textbf{0.807}  
\end{tabular}

\label{tab_ablation}
\end{table}

\begin{table}[!htb]
\centering
\caption{Hyper-parameters investigation of the neighbor number $K$ and the number of sampled contour points $n$ on the CoNSeP dataset. 
}
\begin{tabular}{c|cccccc}
\cline{1-6}
n=18 & $F_{avg}$ & {$F^m_{c}$}  & {$F^i_c$}  & {$F^e_c$}  & {$F^s_c$}\\ \hline 
K=3         & 0.752 & 0.785 & 0.577  & 0.915 & 0.734   \\
K=4              & 0.777 & 0.726 & 0.685  & 0.890 & 0.807   \\ 
K=6  & 0.727 & 0.721 & 0.591 & 0.895 & 0.704   \\\hline
\multirow{1}{*}{K=4}  
& $F_{avg}$ & {$F^m_{c}$}  & {$F^i_c$}  & {$F^e_c$}  & {$F^s_c$}\\ \hline 
n=9         & 0.727 & 0.741 & 0.577  & 0.912 & 0.681   \\
n=18          & 0.777  & 0.726 & 0.685  & 0.890 & 0.807    \\
n=36        & 0.739 & 0.762 & 0.576  & 0.914 & 0.705   \\
\end{tabular}

\label{tab_hyper}
\end{table}
\subsection{Ablation study}
We perform the ablation study on the CoNSeP dataset and the 'Inst-gt' setting. In TABLE~\ref{tab_ablation}, (a)`Baseline' consists of the pixel-wise feature extraction branch and a simple nuclei feature extraction module without PSL and GSL. (b)`+GCN' is the baseline using a GCN without edge features. (c)`+EdgeFeat' denotes the baseline using a GCN with edge features. (d)`Ours' is the proposed framework with the PSL \& GSL modules.
Comparing (a) to (b), using the GCN significantly improves the classification performance by around 21.4\% average F-score. That shows the powerful ability of the graph network in modeling relationships. Comparing (c) to (b) shows that the proposed edge features improve the average F-score by 3.9\%. It indicates that the background information provided by the proposed edge feature helps better identify the nuclei types. Comparing (d) to (c) suggests that our proposed PSL module leads to a great increase of 8.7\% average F-score. Overall, the proposed framework surpasses the baseline by 34.0\% average F-score when the segmentation GT is available. 

\textbf{Computational efficiency} 
We evaluate our proposed classification framework on a machine with Ubuntu 20.04, a NVIDIA-A6000 GPU with 48 GB memory, and Intel(R) Xeon(R) W-2235 CPU with 64 GB memory. Training our proposed method cost 8 hours to 2 days for these four datasets. The GPU memory cost and inference time for an image patch of size $1000 \times 1000$ are about 6 GB and 1.97s. The size of the classification model is around 300 MB. 

\subsection{Investigation of Hyper-parameters}
In TABLE~\ref{tab_hyper}, we study the selection of two hyper-parameters on the CoNSep dataset. One is the neighbor number $K$, which determines how many neighbors are connected to a nucleus in the graph. The other one is the number of sampled contour points $n$, which determines how fine-grained the nucleus shape is in the PSL module.  All the experiments are on the `Inst-gt' setting. 
As TABLE~\ref{tab_hyper} shows, setting $K$ to 4 achieves the highest $F_{avg}$ while setting $K$ to 3/6 results in a drop of 2.5\%/5.0\% F-score. Larger $K$ means more connected neighbors and larger weights for the contextual features. It may be due to that too much context information makes the model pay less attention to the original texture or shape features of predicted nuclei. Thus, setting $K$ to a moderate value mostly benefits our method. Setting $n$ to 18 obtains the best $F_{avg}$ while setting $n$ to 36 causes a drop of 3.8\% $F_{avg}$. It may because the newly sampled points are not distinct enough and could act as noises to cause overfitting. 

\section{Conclusion}
In this paper, we aim to solve a challenging task of automatic nuclei classification for H\&E stained multi-organ histopathology images. First, we propose a novel structure-embedded nuclei classification framework. Second, we build an inter-nuclei graph structure learning module to capture rich contextual information and short-long range correlations among nuclei. Third, we develop an intra-nuclei polygon structure learning module for harvesting better shape representations of a nucleus using a recurrent neural network. The experimental results suggest that both our overall framework and the proposed modules can significantly surpass the existing methods. In the future, it would be meaningful to extend our framework to a unified graph-based nuclei detection and classification model and apply the model to more cancer types of various organs.

\begin{bibliographystyle}{IEEEtran}
\begin{bibliography}{IEEEabrv,tmi}

\end{bibliography}
\end{bibliographystyle}

\end{document}